\documentclass[fleqn,twoside]{article}
\usepackage{espcrc2}

\usepackage{amsmath}
\usepackage{amssymb}
\usepackage{graphicx}
\usepackage{subfigure}

\def\R{\mbox{I\hspace{-.15em}R}}

\title{Time Series Forecasting: Obtaining Long Term Trends with Self-Organizing Maps}

\author{G. Simon\address[GS]{Machine Learning Group - DICE - Universit\'e catholique de Louvain\\
	Place du Levant 3, B-1348 Louvain-la-Neuve, Belgium}
        \thanks{G. Simon is funded by the Belgian F.R.I.A.},
        A. Lendasse\address[AL]{Helsinki University of Technology - Laboratory of Computer and Information Science\\
	Neural Networks Research Centre\\
	P.O. Box 5400, FIN-02015 HUT, FINLAND},
        M. Cottrell\address[MC]{SAMOS-MATISSE, UMR CNRS 8595, Universit\'e Paris I - Panth\'eon Sorbonne\\
	Rue de Tolbiac 90, F-75634 Paris Cedex 13, France},
	J.-C. Fort\address[JCF]{Lab. Statistiques et Probabilit\'es, CNRS C55830, Universit\'e Paul Sabatier Toulouse 3
	Route de Narbonne 118, F-31062 Toulouse Cedex, France}\addressmark[MC]
        and
        M. Verleysen\addressmark[GS]\addressmark[MC]\thanks{M. Verleysen is Senior Research Associate of the Belgian F.N.R.S.}}

\runtitle{2-column format camera-ready paper in \LaTeX}
\runauthor{G. Simon}

\begin{document}

\begin{abstract}
Kohonen self-organisation maps are a well know classification tool, commonly used 
in a wide variety of problems, but with limited applications in time series 
forecasting context. In this paper, we propose a forecasting method specifically 
designed for multi-dimensional long-term trends prediction, with a double application 
of the Kohonen algorithm. Practical applications of the method are also presented.
\vspace{1pc}
\end{abstract}

\maketitle

\section{Introduction}
Time series forecasting is a problem encountered in many fields of applications, 
as finance (returns, stock markets), hydrology (river floods), engineering (electrical 
consumption), etc. Many methods designed for time series forecasting perform well 
(depending on the complexity of the problem) on a rather short-term horizon but are
rather poor on a longer-term one. This is due to the fact that these methods are 
usually designed to optimize the performance at short term, their use at longer 
term being not optimized. Furthermore, they generally carry out the prediction of
a single value while the real problem sometimes requires predicting a vector of 
future values in one step. For example, in the case of some a priori known periodicity, 
it could be interesting to predict all values for a period as a whole. But forecasting 
a vector requires either more complex models (with potential loss of performance for
some of the vector components) or many distinct single value predicting models 
(with potential loss of the correlation information between the various values). 
Methods able to forecast a whole vector with the same precision for each of its 
components are thus of great interest. 

While enlarging the prediction horizon is of course of primary interest for 
practitioners, there is of course some limit to the accuracy that can be expected 
for a long-term forecast. The limitation is due to the availability of the information 
itself, and not to possible limitations of the forecasting methods. Indeed, there is no 
doubt that, whatever forecasting method is used, predicting at long term (i.e. many time 
steps in advance) is more difficult that predicting at short term, because of the missing 
information in the unknown future time steps (those between the last known value and 
the one to predict). At some term, all prediction methods will thus fail. The purpose 
of the method presented in this paper is not to enlarge the time horizon for which 
accurate predictions could be expected, but rather to enlarge the horizon for which 
we can have insights about the future evolution of the series. By insights, we mean 
some information of interest to the practitioner, even if it does not mean accurate 
predictions. For example, are there bounds on the future values ? What can we 
expect in average ? Are confidence intervals on future values large or narrow ?

Predicting many steps in advance could be realized in a straightforward way, by subsampling
the known sequence, then using any short-term prediction method. However, in this case, the
loss of information (used for the forecast) is obviously even higher, due to the lower 
resolution of the known sequence. Furthermore, such solution does not allow in a general 
way to introduce a stochastic aspect to the method, which is a key issue in the proposed 
method. Indeed, to get insights about the future evolution of a series through some 
statistics (expected mean, variance, confidence intervals, quartiles, etc.), several
predictions should be made in order to extract such statistics. The predictions should 
differ; a stochastic prediction method is able to generate several forecasts by repeated 
Monte-Carlo runs. In the method presented in this paper, the stochastic character of the 
method results from the use of random draws on a probability law. 

Another attractive aspect of the method presented in this paper is that it can be used 
to predict scalar values or vectors, with the same expected precision for each component 
in the case of vector prediction. Having at disposal a time series of values $x(t)$ with 
$1 \leq t \leq n$, the prediction of a vector can be defined as follows :
\begin{equation}\label{prob_def}
[x(t+1), \ldots, x(t+d)] = f(x(t), \ldots, x(t-p+1)) + \varepsilon_t
\end{equation}
where $d$ is the size of the vector to be predicted, $f$ is the data generating process, 
$p$ is the number of past values that influence the future values and $\varepsilon_t$ 
is a centred noise vector. The past values are gathered in a $p$-dimensional vector called
{\it regressor}.

The knowledge of $n$ values of the time series (with $n >> p$ and $n >> d$) means that 
relation (\ref{prob_def}) is known for many ($n-p-d+1$) time steps in the past. The 
modeling problem then becomes to estimate a function $f$ that models correctly the
time series for the whole set of past regressors. 

The idea of the method is to segment the space of $p$-dimensional regressors. 
This segmentation can be seen as a way to make possible a local modeling in each segment. 
This part of the method is achieved using the Self-Organizing Map (SOM) \cite{Kohonen95}. 
The prototypes obtained for each class model locally the regressors of the corresponding 
class. Furthermore, in order to take into account temporal dependences in the series, 
deformation regressors are built. Those vectors are constructed as the differences between 
two consecutive regressors. The set of regressor deformations can also be segmented using 
the SOM. Once those two spaces are segmented and their dependences characterized, simulations 
can be performed. Using a kind of Monte-Carlo procedure to repeat the simulations, it is 
then possible to estimate the distribution of these simulations and to forecast global
trends of the time series at long term.

Though we could have chosen some other classical vector quantization (VQ) method as only 
the clustering property is of interest here, the choice of the SOM tool to perform 
the segmentation  of the two spaces is justified by the fact that SOM are efficient 
and fast compared to other VQ methods with a limited complexity \cite{deBodt04} and 
that they provide an intuitive and helpful graphical representation. 

In the following of this paper, we first recall some basic concepts about the SOM 
classification tool. Then we introduce the proposed forecasting method, the double 
vector quantization, for scalar time series and then for vector ones. Next we present 
some experimental results for both scalar and vector forecastings. A proof of the method 
stability is given in appendix.

\section{The Kohonen Self-Organizing Maps}
The Self-Organizing Maps (SOM), developed by Teuvo Kohonen in the 80's \cite{Kohonen95}, 
has now become a well-known tool, with established properties \cite{Cottrell98}, 
\cite{Cottrell97}. Self-Organizing Maps have been commonly used since their first 
description in a wide variety of problems, as classification, feature extraction, 
pattern recognition and other related applications. As shown in a few previous works
\cite{Cottrell96}, \cite{Cottrell98b}, \cite{Walter90}, \cite{Vesanto97}, \cite{Koskela98},
\cite{Lendasse98}, the SOM may also be used to forecast time series at short term.

The Kohonen Self-Organizing Maps (SOM) can be defined as an unsupervised classification
algorithm from the artificial neural network paradigm. Any run of this algorithm results 
in a set, with a priori fixed size, of prototypes. Each one of those prototypes 
is a vector of the same dimension as the input space. Furthermore, physical
neighbourhood relation links the prototypes.  Due to this neighbourhood relation, 
we can easily graphically represent the prototypes in a 1- or 2-dimensional grid.

After the learning stage each prototype represents a subset of the initial input set 
in which the inputs share some similar features. Using Voronoi's terminology, the prototype 
corresponds to a centroid of a region or zone, each zone being one of the classes 
obtained by the algorithm. The SOM thus realizes a vector quantization of the input space 
(a Voronoi tessellation) that respects the original distribution of the inputs. Furthermore,
a second property of the SOM is that the resulting prototypes are ordered according to
their location in the input space. Similar vectors in the input space are associated either 
to the same prototype (as in classical VQ) or to two prototypes that are neighbours 
on the grid. This last property, known as the topology preservation, does not hold for other 
standard vector quantization methods like competitive learning.

The ordered prototypes of a SOM can easily be represented graphically, allowing a more 
intuitive interpretation: the 1- or 2-dimensional grid can be viewed as a 1- or 2-dimensional 
space where the inputs are projected by the SOM algorithm, even if, in fact, the inputs 
are rather projected on the prototypes themselves (with some interpolation if needed 
in the continuous case). This projection operation for some specific input is proceeded
by determining the nearest prototype with respect to some distance metric (usually the 
Euclidian distance).

\section{The double quantization method}\label{descr}
The method described here aims to forecast long-term trends for a time series evolution. 
It is based on the SOM algorithm and can be divided into two stages: the characterization 
and the forecasting. The characterization stage can be viewed as the learning, while the 
forecasting can be viewed as the use of a model in a generalization procedure.

For the sake of simplicity, the method is first presented for scalar time series prediction 
(i.e. $d = 1$ in (\ref{prob_def})) and then detailed later on for vector forecasting. 
Examples of the method application to scalar and vector time series will be provided 
in section \ref{expe}.

\subsection{Method description: characterization}
Though the determination of an optimal regressor in time series forecasting (at least 
in a nonlinear prediction case) is an interesting and open question \cite{Verleysen99}, 
it is considered here that the optimal, or at least an adequate, regressor of the time
series is known. Classically, the regressor can for example be chosen according to some 
statistical resampling (cross-validation, bootstrap, etc.) procedure.

As for many other time series analysis methods, conversion of the inputs into regressors 
leads to $n-p+1$ vectors in a $p$-dimensional space, where $p$ is the regressor size and 
$n$ the number of values at our disposal in the time series. The resulting
regressors are denoted:
\begin{equation}\label{regress}
x^t_{t-p+1} = \{x(t), x(t-1), \ldots, x(t-p+1)\},
\end{equation}
where $p \leq t \leq n$, and $x(t)$ is the original time series at our disposal with 
$1 \leq t \leq n$. In the above $x^t_{t-p+1}$ notation, the subscript index denotes the
first temporal value of the vector, while the superscript index denotes its last 
temporal value.

The obtained vectors $x^t_{t-p+1}$ are then manipulated and the so-called deformations 
$y^t_{t-p+1}$ are created according to:
\begin{equation}\label{deforms}
y^t_{t-p+1} = x^{t+1}_{t-p+2} - x^t_{t-p+1}.
\end{equation}
Note that, by definition, each $y^t_{t-p+1}$ is associated to one of the $x^t_{t-p+1}$. 
In order to highlight this link, the same indices have been used.

Putting all $y^t_{t-p+1}$ together in chronological order forms another time series 
of vectors, the deformations series in the so-called deformation space to be opposed 
to the original space containing the regressors $x^t_{t-p+1}$. Of course, there exist 
$n-p$ deformations of dimension $p$. 

The SOM algorithm can then be applied to each one of these two spaces, quantizing both 
the original regressors $x^t_{t-p+1}$ and the deformations $y^t_{t-p+1}$ respectively. 
Note that in practice any kind of SOM map can be used, but it is assumed that 
one-dimensional maps (or strings) are more adequate in this context.

As a result of the vector quantization by the SOM on all $x^t_{t-p+1}$ of the original space, 
$n_1$ $p$-dimensional prototypes $\bar{x}_i$ are obtained ($1 \leq i \leq n_1$). The clusters 
associated to $\bar{x}_i$ are denoted $c_i$. The second application of the SOM on all 
deformations $y^t_{t-p+1}$ in the deformation space results in $n_2$ $p$-dimensional prototypes 
$\bar{y}_j$, $1 \leq j \leq n_2$. Similarly the associated clusters are denoted $c'_j$.

To perform the forecasting, more information is needed than the two sets of prototypes. We
therefore compute a matrix $f(ij)$ based on the relations between the $x^t_{t-p+1}$ and 
the $y^t_{t-p+1}$ with respect to their clusters ($c_i$ and $c'_j$ respectively). The row 
$f_{ij}$ for a fixed $i$ and $1 \leq j \leq n_2$ is the conditional probability that 
$y^t_{t-p+1}$ belongs to $c'_j$, given that $x^t_{t-p+1}$ belongs to $c_i$. In practice, those
probabilities are estimated by the empirical frequencies:
\begin{equation}\label{freqs}
f_{ij} = \frac{\#\{x^t_{t-p+1} \in c_i\ {\rm and} \ y^t_{t-p+1} \in c'_j\}}{\#\{x^t_{t-p+1} \in c_i\}}
\end{equation}
with $1 \leq i \leq n_1$, $1 \leq j \leq n_2$.

Note that, for a fixed $i$, elements $f_{ij}$ ($1 \leq j \leq n_2$) sum to one; 
this justifies the fact that each row of the matrix is an (empirically estimated)
probability law. Therefore the matrix will be called {\it transition matrix} in the following.

The computation of this transition matrix completes the characterization part of the method.

\subsection{Method description: forecasting}
Once the prototypes in the original and deformation spaces together with the transition 
matrix are known, we can forecast a time series evolution over a rather long-term horizon $h$ 
(where horizon 1 is defined as the next value $t + 1$ for time $t$).

The methodology for such forecasting can be described as follows. First, consider a time value
$x(t)$ for some time $t$. The corresponding regressor is $x^t_{t-p+1}$. Therefore we can find 
the associated prototype in the original space, for example $\bar{x}_k$ (this operation 
is in fact equivalent to determining the class $c_k$ of $x^t_{t-p+1}$ in the SOM). We then
look at row $k$ in the transition matrix and randomly choose a deformation prototype $\bar{y}_l$ 
among the $\bar{y}_j$ according to the conditional probability distribution defined by 
$f_{kj}$, $1 \leq j \leq n_2$. The prediction for time $t+1$ is obtained according to
relation (\ref{deforms}):
\begin{equation}\label{preds}
\hat{x}^{t+1}_{t-p+2} = x^t_{t-p+1} + \bar{y}_l,
\end{equation}
where $\hat{x}^{t+1}_{t-p+2}$ is the estimate of the true $x^{t+1}_{t-p+2}$ given by our 
time series prediction model. However $\hat{x}^{t+1}_{t-p+2}$ is in fact a $p$-dimensional 
vector, with components corresponding to times from $t-p+2$ to $t+1$ (see relations 
(\ref{regress}) and (\ref{deforms})). As in the scalar case considered here we are only 
interested in a single estimate at time $t+1$, we extract the scalar prediction $\hat{x}(t+1)$ 
from the $p$-dimensional vector $\hat{x}^{t+1}_{t-p+2}$. 

We can iterate the described procedure, plugging in $\hat{x}(t+1)$ for $x(t)$ in 
(\ref{regress}) to compute $\hat{x}^{t+2}_{t-p+3}$ by (\ref{preds}) and extracting 
$\hat{x}(t+2)$. We then do the same for $\hat{x}(t+3)$, $\hat{x}(t+4)$, \ldots, 
$\hat{x}(t+h)$. This ends the run of the algorithm to obtain a single simulation 
of the series at horizon $h$.

Next, as the goal of the method is not to perform a single long-term simulation, 
the simulations are repeated to extract trends. Therefore a Monte-Carlo procedure 
is used to repeat many times the whole long-term simulation procedure at horizon $h$, 
as detailed above. As part of the method (random choice of the deformation according
to the conditional probability distributions given by the rows of the transition matrix)
is stochastic, repeating the procedure leads to different simulations. Observing those 
evolutions allows estimating the simulation distribution and infer global trends 
of the time series, as the evolution of its mean, its variance, confidence intervals, etc.

It should be emphasized once again that the double quantization method is not designed 
to determine a precise estimate for time $t+1$ but is more specifically devoted 
to the problem of longterm evolution, which can only be obtained in terms of trends.

\subsection{Generalisation: vector forecasting}
Suppose that it is expected to predict vectors $x^{t+d}_{t+1}$ of future values of the times 
series $x(t)$; $x^{t+d}_{t+1}$ is a vector defined as:
\begin{equation}\label{vect_prob}
x^{t+d}_{t+1} = \{x(t+d), \ldots, x(t+2), x(t+1)\},
\end{equation}
where $d$ is determined according to a priori knowledge about the series. For example 
when forecasting an electrical consumption, it could be advantageous to predict all hourly 
values for one day in a single step instead of predicting iteratively each value separately.

As above regressors of this kind of time series can be constructed according to:
\begin{equation}\label{vect_regress}
x^t_{t-p+1} = \{x^t_{t-d+1}, x^{t-d}_{t-2d+1}, \ldots, x^{t-p+d}_{t-p+1}\},
\end{equation}
where $p$, for the sake of simplicity, is supposed to be a multiple of $d$ though 
this is not compulsory. The regressor $x^t_{t-p+1}$ is thus constructed as the concatenation 
of $d$-dimensional vectors from the past of the time series, as it is the concatenation
of single past values in the scalar case. As the $x^t_{t-p+1}$ regressor is composed of $p/d$ 
vectors of dimension $d$, $x^t_{t-p+1}$ is a $p$-dimensional vector. 

Deformation can be formed here according to:
\begin{equation}\label{vect_deforms}
y^t_{t-p+1} = x^{t+d}_{t-p+d+1} - x^t_{t-p+1}.
\end{equation}

Here again, the SOM algorithm can be applied on both spaces, classifying both the regressors
$x^t_{t-p+1}$ and the deformations $y^t_{t-p+1}$ respectively. We then have $n_1$ prototypes 
$\bar{x}_i$ in the original space, with $1 \leq i \leq n_1$, associated to classes $c_i$.
In the deformation space, we have $n_2$ prototypes $\bar{y}_j$, $1 \leq j \leq n_2$, associated 
to classes $c'_j$. 

A transition matrix can be constructed as a vector generalisation of relation (\ref{freqs}):
\begin{equation}\label{vect_freqs}
f_{ij} = \frac{\#\{x^t_{t-p+1} \in c_i \ {\rm and} \ y^t_{t-p+1} \in c'_j\}}{\#\{x^t_{t-p+1} \in c_i\}}
\end{equation}
with $1 \leq i \leq n_1$, $1 \leq j \leq n_2$.

The simulation forecasting procedure can also be generalised:
\begin{itemize}
\item consider the vector input $x^t_{t-d+1}$ for time $t$. The corresponding regressor 
is $x^t_{t-p+1}$;
\item find the corresponding prototype $\bar{x}_k$;
\item choose a deformation prototype $\bar{y}_l$ among the $\bar{y}_j$ 
according to the conditional distribution given by elements $f_{kj}$ of row $k$;
\item forecast $\hat{x}^{t+d}_{t-p+d+1}$ as 
\begin{equation}\label{vect_pred}
\hat{x}^{t+d}_{t-p+d+1} = x^t_{t-p+1} + \bar{y}_l;
\end{equation}
\item extract the vector
$$\{\hat{x}(t+1), \hat{x}(t+2), \ldots, \hat{x}(t+d)\}$$
from the $d$ first columns of $\hat{x}^{t+d}_{t-p+d+1}$;
\item repeat until horizon $h$.
\end{itemize}

For this vector case too, a Monte-Carlo procedure is used to repeat many times the whole 
longterm simulation procedure at horizon $h$. Then the simulation distribution and its 
statistics can be observed. This information gives trends for the long term of the 
time series. 

Note that using the SOM to quantize the vectors $x^t_{t-p+1}$ and $y^t_{t-p+1}$, 
the method reaches the goal of forecasting vectors with the same precision for each 
of their components. Indeed each component from regressors $x^t_{t-p+1}$ and $y^t_{t-p+1}$ 
has the same relative weight while the distance between the considered regressor 
and prototype is computed in the SOM algorithm. None of the $x^t_{t-p+1}$ or $y^t_{t-p+1}$ 
components have thus a greater importance in the modification of the prototype weight
during the learning of the SOM.

\subsection{Extensions}
Two important comments must be done.

First, as illustrated in both examples below, it is not mandatory (in equations (\ref{prob_def}), 
(\ref{regress}), (\ref{vect_prob}), (\ref{vect_regress})) to consider all successive values 
in the regressor; according to the knowledge of the series or to some validation procedure, 
it might be interesting to select regressors with adequate, but not necessarily successive, 
scalar values or vectors in the past. 

Secondly, the vector case has been illustrated in the previous section on temporal vectors 
(see equation (\ref{vect_prob})). An immediate extension of the method would be to consider 
spatial vectors, for example when several series must be predicted simultaneously.
The equations in the previous section should be modified, but the principle of the
method remains valid.

\subsection{Method stability}
The predictions obtained by the model described in the previous subsections should ideally
be confined in the initial space defined by the learning data set. In that case, the series of
predicted values $y^t_{t-p+1}$ is said to be stable. Otherwise, if the series tends to infinity 
or otherwise diverges, it is said to be unstable. The method has been proven to be stable 
according to this definition; a proof is given in appendix.

\section{Experimental results}\label{expe}
This section is devoted to the application of the method on two times series. The first 
one is the well-known Santa Fe A benchmark presented in \cite{Weigend94}; it is a scalar 
time series. The second time series is the Polish electrical consumption from 1989 to 1996 
\cite{Cottrell98b}. This real-world problem requires the prediction of a vector 
of 24 hourly values.

\subsection{Methodology}
In the method description, the numbers $n_1$ and $n_2$ of prototypes have not been fixed. 
Indeed, the problem is that different values of $n_1$ ($n_2$) result in different 
segmentations in the original (deformation) space and in different conditional
distribution in the transition matrix. The model may thus slightly vary.

Selecting the best values for $n_1$ and $n_2$ is an important question too. Traditionally, 
such hyperparameters are estimated by model selection procedures such as AIC, BIC or 
computationally-costly resampling techniques (Leave-One-Out, k-fold cross validation, bootstrap). 
As it will be shown further in this paper, exact values of $n_1$ and $n_2$ are not necessary, 
as the sensitivity of the method around the optimums is low. A simple validation is 
then used to choose adequate values for $n_1$ and $n_2$. For that purpose the available
data are divided into three subsets: the learning, the validation and the test set. 
The learning set is used to fix the values of the model parameters, such as the weights 
of the prototypes in the SOM and the transition matrix. The validation set is used to fix 
meta-parameters, such as the numbers $n_1$ and $n_2$ of prototypes in the SOM maps. The validation 
set is thus used for model selection. The test set aims to see how the model behaves on unused 
data that mimic real conditions. 

The selection of $n_1$ and $n_2$ is done with regards to an error criterion, in our case 
a sum of squared error criterion, computed over the validation set $VS$:
\begin{equation}\label{SSE}
\displaystyle e_{SSE} = \sum_{y(t+1) \in VS}^{}{(y(t+1)-\hat{y}(t+1))^2}.
\end{equation}

Once $n_1$ and $n_2$ have been chosen, a new learning is done with a new learning set 
obtained from the reassembled learning and validation sets. This new learning is only 
performed once with optimal values for $n_1$ and $n_2$.

Note that, hopefully, the sensitivity of the method to specific values of $n_1$ and $n_2$ 
is not high. This has been experimentally verified in all our simulations, and will be 
illustrated on the first example (Santa Fe A) in section \ref{SFA}.

Another crucial question is the sensitivity of the method to various runs of the SOM 
algorithm (with the same $n_1$ and $n_2$ values). Indeed it is well known that initial 
conditions largely influence the exact final result of the SOM algorithm (by final result 
it is meant the prototype locations, and their neighborhood relations) \cite{deBodt02}.
Nevertheless, as mentioned above, the neighborhood relations of the SOM are used for 
visualization purposes only; they do not influence the results of the forecast. Moreover, 
the location of the centroids are used to quantize the space (therefore allowing the estimation 
of the empirical conditional frequencies of the clusters); small variations in the centroid 
location have thus a low influence on each prediction generated by the method, and an even 
lower one on the statistics (mean, confidence intervals, etc.) estimated from the predictions. 
This last result has been confirmed experimentally in all our simulations, for which 
no significant difference was observed after different runs of the two SOM algorithms.

\subsection{Scalar forecasting: Santa Fe A}\label{SFA}
The Santa Fe A time series \cite{Weigend94} has been obtained from a far-infrared-laser 
in a chaotic state. This time series has become a well-known benchmark in time series 
prediction since the Santa Fe competition in 1991. The completed data set contains
10 000 data. This set has been divided here as follows: the learning set contains 6000 data,
the validation set 2000 data, and test set 100 data. Note that the best neural network models
described in \cite{Weigend94} do not predict much more than 40 data, making a 100-data test 
set a {\it very} long-term forecasting. 

Here, the regressors $x^t_{t-p+1}$ have been constructed according to 
\begin{eqnarray}\label{regress_SFA}
x^t_{t-p+1} & = & \{x(t), x(t-1), x(t-2),\nonumber \\
 & & x(t-3), x(t-5), x(t-6)\}.
\end{eqnarray}
This choice is made according to previous experience on this series \cite{Weigend94}. 
In other words, $d = 1$, $p = 6$ (as value $x(t-4)$ is omitted) and $h = 100$.

\begin{figure}[!hbt]
\begin{center}
\scalebox{0.30}{\includegraphics{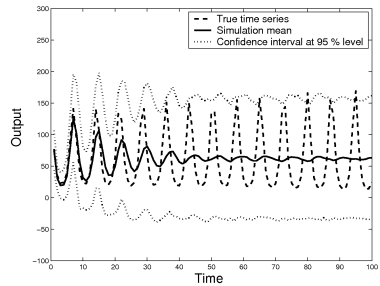}}
\caption{Comparison between the mean of the 1000 simulations (solid) and the true values (dashed), together with confidence intervals at 95\% level (dotted).}
\label{pred_SFA_LT}
\end{center}
\end{figure}

\begin{figure}[!hbt]
\begin{center}
\scalebox{0.30}{\includegraphics{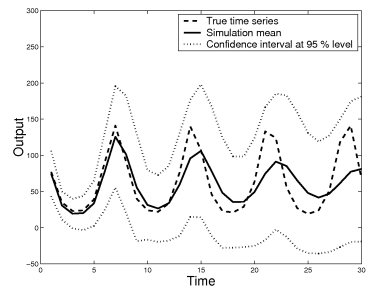}}
\caption{Comparison for the first 30 values between the mean of the 1000 simulations (solid) and the true values of the test set (dashed), together with confidence intervals at 95\% level (dotted).}
\label{pred_SFA_LT_zoom}
\end{center}
\end{figure}

\begin{figure}[!hbt]
\begin{center}
\scalebox{0.30}{\includegraphics{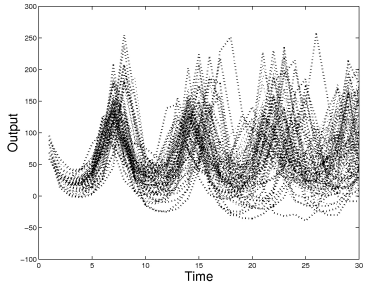}}
\caption{100 simulations picked out at random from the 1000 simulations made for the Santa Fe A long-term forecasting.}
\label{100_simu_SFA}
\end{center}
\end{figure}

\begin{figure}[!hbt]
\begin{center}
\scalebox{0.30}{\includegraphics{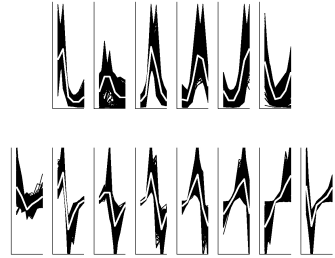}}
\caption{The code vectors and associated curves in the regressor (top) and deformation (bottom) spaces (when $n_1$ = 6 and $n_2$ = 8). 
The code vectors are represented in white as 6-dimensional vectors (according to (\ref{regress_SFA})).  
Regressors (resp. deformations) belonging  to each class are shown in black.}
\label{code_vector_SFA}
\end{center}
\end{figure}

In this simulation, Kohonen strings of 1 up to 200 prototypes in each space have been used. 
All the 40 000 possible models have been tested on the validation set. The best model among 
them has 179 prototypes in the regressor space and 161 prototypes in the deformation space. 
After relearning this model on both the learning and validation sets, 1000 simulations were 
performed on a horizon of 100. Then the mean and confidence interval at 95\% level were computed, 
giving information on the time series trends. Figure \ref{pred_SFA_LT} shows the mean 
of the 1000 simulations compared to the true values contained in the test set, together 
with the confidence interval at 95\% level. Figure \ref{pred_SFA_LT_zoom} shows a zoom 
on the first 30 values. In figure \ref{100_simu_SFA}, we can see 100 simulations 
for the same 30 values. Note the stability obtained through the replications. 
For a simpler model with $n_1 = 6$ and $n_2 = 8$ (used for illustrations purposes), 
figure \ref{code_vector_SFA} shows the code vectors and regressors (resp. deformations) 
in each class; table \ref{trans_matrix_SFA} shows the corresponding transition matrix.

From figure \ref{pred_SFA_LT_zoom}, it should be noted that the method gives roughly 
the first 25 values of the time series, a result that is not so far from those
obtained with the best neural network models of the Santa Fe competition \cite{Weigend94}.

From figure \ref{pred_SFA_LT}, we can infer that the series mean will neither increase 
nor decrease. In addition, the confidence interval does contain the whole evolution 
of the time series for the considered 100 future values. The trend for long term forecasting
is thus that the series, though chaotic, will show some kind of stability in its evolution 
for the next 100 values.

\begin{figure}[!hbt]
\begin{center}
\scalebox{0.30}{\includegraphics{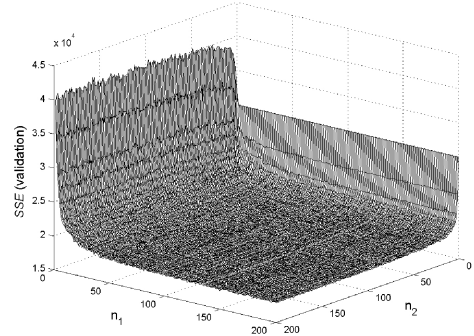}}
\caption{Impact of the variation of $n_1$ and $n_2$ on the model generalization ability for the Santa Fe A time series.}
\label{SSE_SFA}
\end{center}
\end{figure}

As all the 40 000 models have been generated and learned, the influence of varying the 
$n_1$ and $n_2$ values can be observed. This influence is illustrated in figure \ref{SSE_SFA}. 
It is clear from this figure that there is a large flat region around the optimal values; 
in this region, all models generalize rather equivalently. This justifies, a posteriori,
the choice of a simple resampling method to choose $n_1$ and $n_2$.

\begin{table}[!htb]
{\footnotesize
\begin{tabular}{|c|c|c|c|c|c|c|c|}
\hline
0.12 & 0 & 0 & 0 & 0 & 0 & 0.23 & 0.66\\
\hline
0.67 & 0.30 & 0 & 0 & 0 & 0 & 0.02 & 0.01\\
\hline
0.05 & 0.55 & 0.40 & 0 & 0 & 0 & 0 & 0\\
\hline
0.03 & 0 & 0.30 & 0.54 & 0.13 & 0 & 0 & 0\\
\hline
0 & 0 & 0 & 0 & 0.50 & 0.48 & 0.02 & 0\\
\hline
0.06 & 0 & 0 & 0 & 0 & 0.34 & 0.56 & 0.04\\
\hline
\end{tabular}
}
\caption{Example of transition matrix, here with $n_1 = 6$
and $n_2 = 8$ as in figure \ref{code_vector_SFA}. Note that in each row,
the frequency values sum to one.}
\label{trans_matrix_SFA}
\end{table}

\subsection{Vector forecasting: the Polish electrical consumption}\label{Polish}
\begin{figure}[!hbt]
\begin{center}
\scalebox{0.30}{\includegraphics{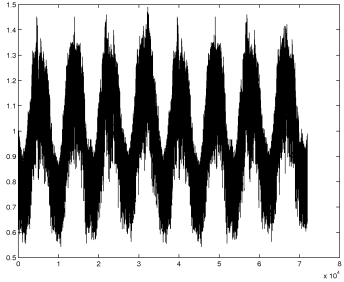}}
\caption{The Polish electrical consumption time series, between 1989 and 1996.}
\label{series_Polish}
\end{center}
\end{figure}

As second example, we use the Polish electrical load time series \cite{Cottrell98b}. 
This series contains hourly values from 1989 to 1996. The whole dataset contains
about 72 000 hourly data and is plotted in figure \ref{series_Polish}. 
Due to the daily periodicity of the time series, we are interested in daily predictions. 
This is thus an illustration of the case $d > 1$, since it seems natural to forecast 
the 24 next values in one step (the next day), the time window becoming daily instead of hourly.

Having now at our disposal 3000 $x^t_{t-p+1}$ data of dimension 24, we use 2000 of them 
for the learning, 800 for a simple validation and 200 for the test. Since the optimal 
regressor is unknown, many different regressors were tried, using intuitive understanding 
of the process. The final regressor is:
\begin{eqnarray}\label{regress_Polish}
x^t_{t-p+1} & = & \{x^t_{t-24+1}, x^{t-24}_{t-48+1}, x^{t-48}_{t-72+1}, \nonumber \\
 & & \hspace{0.5cm} x^{t-144}_{t-168+1}, x^{t-168}_{t-192+1}\},
\end{eqnarray}
that is the 24 hourly values of today, of yesterday, of two, six and seven days ago. 
This regressor is maybe not the optimal one, but it is the one that makes the lowest 
error on the validation set in comparison with other tested ones. Since the regressor 
contains $p = 5$ data of dimension $d = 24$, we work in a 120-dimensional space. 
We then run the algorithm again on the learning set with values for $n_1$ and $n_2$ 
each varying from 5 to 200 prototypes by steps of 5. The lowest error is made by a model 
with $n_1 = 160$ and $n_2 = 140$ respectively.

\begin{figure}[!hbt]
\begin{center}
\scalebox{0.30}{\includegraphics{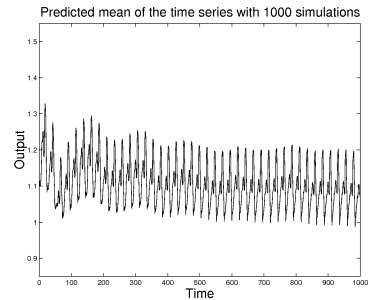}}
\caption{Mean of the 1000 simulations at long term ($h = 40$).}
\label{pred_Polish_LT}
\end{center}
\end{figure}

\begin{figure}[!hbt]
\begin{center}
\scalebox{0.30}{\includegraphics{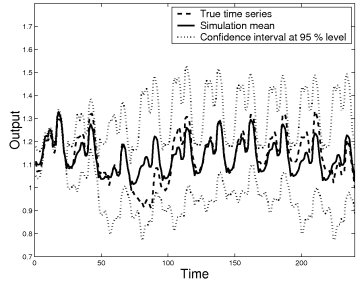}}
\caption{Comparison between the true values (dashed), the mean of the predictions (solid) and the confidence interval at 95 \% level (dotted).}
\label{pred_Polish_zoom}
\end{center}
\end{figure}

Another model is then learned with 160 and 140 parameter vectors in each space with the new
learning set, now containing 2000+800 data. The forecasting obtained from this model is repeated
1000 times. Figure \ref{pred_Polish_LT} presents the mean of the 1000 simulations obtained 
with 24-dimensional vectors and with horizon $h$ limited to 40 days (a single plot of the 
whole 24 * 200 predicted values becomes unreadable). For convenience, figure \ref{pred_Polish_zoom}
shows a zoom and a comparison between the mean of those 1000 long-term predictions and the
real values. A confidence interval at 95\% level is also provided.

From figure \ref{pred_Polish_zoom}, it is clear that the mean of the prediction at long term 
will show the same periodicity as the true time series and that the values will be contained 
in a rather narrow confidence interval. This fact denotes a probable low variation of the series 
at long term. 

\begin{figure}[!hbt]
\begin{center}
\scalebox{0.30}{\includegraphics{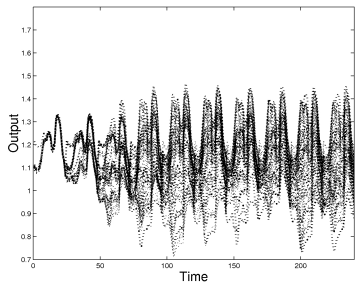}}
\caption{Plot of 100 simulations chosen at random from the 1000 simulations.}
\label{100_simu_Polish}
\end{center}
\end{figure}

Figure \ref{100_simu_Polish} shows 100 predictions obtained by the Monte-Carlo procedure 
picked up at random before taking the mean. See that different simulations have about 
the same shape; this is a main argument for determining long-term trends.

\begin{figure}[!hbt]
\begin{center}
\scalebox{0.30}{\includegraphics{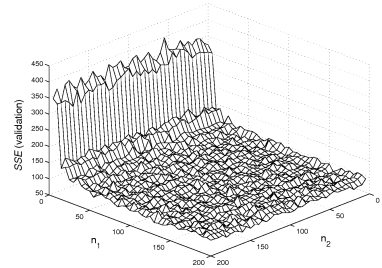}}
\caption{Impact of the variation of $n_1$ and $n_2$ on the model generalization ability for the Polish electrical consumption problem.}
\label{SSE_Polish}
\end{center}
\end{figure}

Finally, as in the previous example, the influence of $n_1$ and $n_2$ can be observed. 
In figure \ref{SSE_Polish}, a very large flat region is also present around the
best model. Sub-optimal selection of the $n_1$ and $n_2$ values will thus not penalize 
too heavily the model generalization abilities.

\section{Conclusion}
In this paper, we have presented a time series forecasting method based on a double 
classification of the regressors and of their deformations using the SOM algorithm. 
The use of SOMs makes it possible to apply the method both on scalar and vector time 
series, as discussed in section \ref{descr} and illustrated in section \ref{expe}. 
A proof of the method stability is given in appendix.

The proposed method is not designed to obtain an accurate forecast of the next values 
of a series, but rather aims to determine long-term trends. Indeed, its stochastic 
nature allows repeating simulations by a Monte-Carlo procedure, allowing to compute 
statistics (variance, confidence intervals, etc.) on the predictions. Such a method could
also be used for example in the financial context, for the estimation of volatilities.

\section*{Acknowledgements}
We would like to thank Professor Osowsky from Warsaw Technical University for providing 
us the Polish Electrical Consumption data used in our example.

\section*{Appendix}
\section*{Method stability}
Intuitively, the stability property of the method is not surprising. Indeed, the model 
is designed such that it will mostly produce predictions that are in the range of the observed 
data. By construction, deformations are chosen randomly according to an empirical probability 
law and the obtained predictions should stay in the same range. If, for some reason, 
the prediction is about to exceed this range during one of the simulations, the next deformations 
will then tend to drive it back inside this range, at least with high probability. Furthermore, 
as simulations are repeated with the Monte-Carlo procedure, the influence of such unexpected 
cases will be reduced when the mean is taken to obtain the final predictions. The following 
of this section is intended to prove this intuitive result.

The proof consists in two steps: it is first shown that the series generated by the model
is a Markov chain; secondly, it is demonstrated that this particular type of Markov chain 
is stable. In order to improve the readability of the proof, lighter notations will be used. 
For a fixed $d$ and a fixed $p$, notation $X_t$ will represent the vector $x^t_{t-p+1}$. 
The last known regressor will be denoted $X_0$. The prototype of a cluster $C'_j$ of
deformations will be noted $Y_j$ . Finally, hats will be omitted for simplicity as 
all regressors $X_t$ are estimations, except for $t = 0$.

To prove that the series is a Markov chain, we consider the starting vector of the simulation 
at time 0. The corresponding initial regressor of the series is denoted $X_0$, and $C_0$ 
is the corresponding SOM cluster in the regressor space. The deformation that is applied 
to $X_0$ at this stage is $Y_0$. Then the next values of the series are given by 
$X_1 = X_0 + Y_0$, $X_2 = X_0 + Y_0 + Y_1$, \ldots, with $Y_0$, $Y_1$, \ldots drawn randomly 
from the transition matrix for clusters $C_0$, $C_1$, \ldots respectively. The series $X_t$ 
is therefore a Markov chain, homogeneous in time (the transition distribution are not
time dependant), irreducible and defined over a numerable set (the initial $X_t$ are in 
finite number, and so are the deformations).

To show the stability of this Markov chain and thus the existence of a stationary distribution,
Foster's criterion \cite{Fayolle95} is applied. Note that this criterion is a stronger 
result which proves the ergodicity of the chain, which in turns implies the stability. 
Foster's criterion is the following: 

A necessary and sufficient condition for an irreducible chain to be ergodic is 
that there exists a positive function $g(.)$, a positive $\varepsilon$ and a finite
set $A$ such that:
\begin{equation}\label{Foster}
\begin{array}{l}
\forall x \in \Omega : E(g(X_{t+1}) | X_t = x) < \infty, \\
\forall x \notin \Omega : E(g(X_{t+1}) | X_t = x) - g(x) \leq - \varepsilon.
\end{array}
\end{equation}

Since the Markov chain is homogenous, it is sufficient to observe transition $Y_0$ 
from $X_0$ to $X_1$. The same development can be deduced for any other transition.

The demonstration is done for two-dimensional regressors but can be generalized easily 
to other dimensions. Note that in the following, we use $g(.) = \|.\|^2$ in (\ref{Foster}).

\begin{figure}[!hbt]
\begin{center}
\subfigure[A Cluster within an acute angle]{\label{Prop_1a}\scalebox{0.2}{\includegraphics*{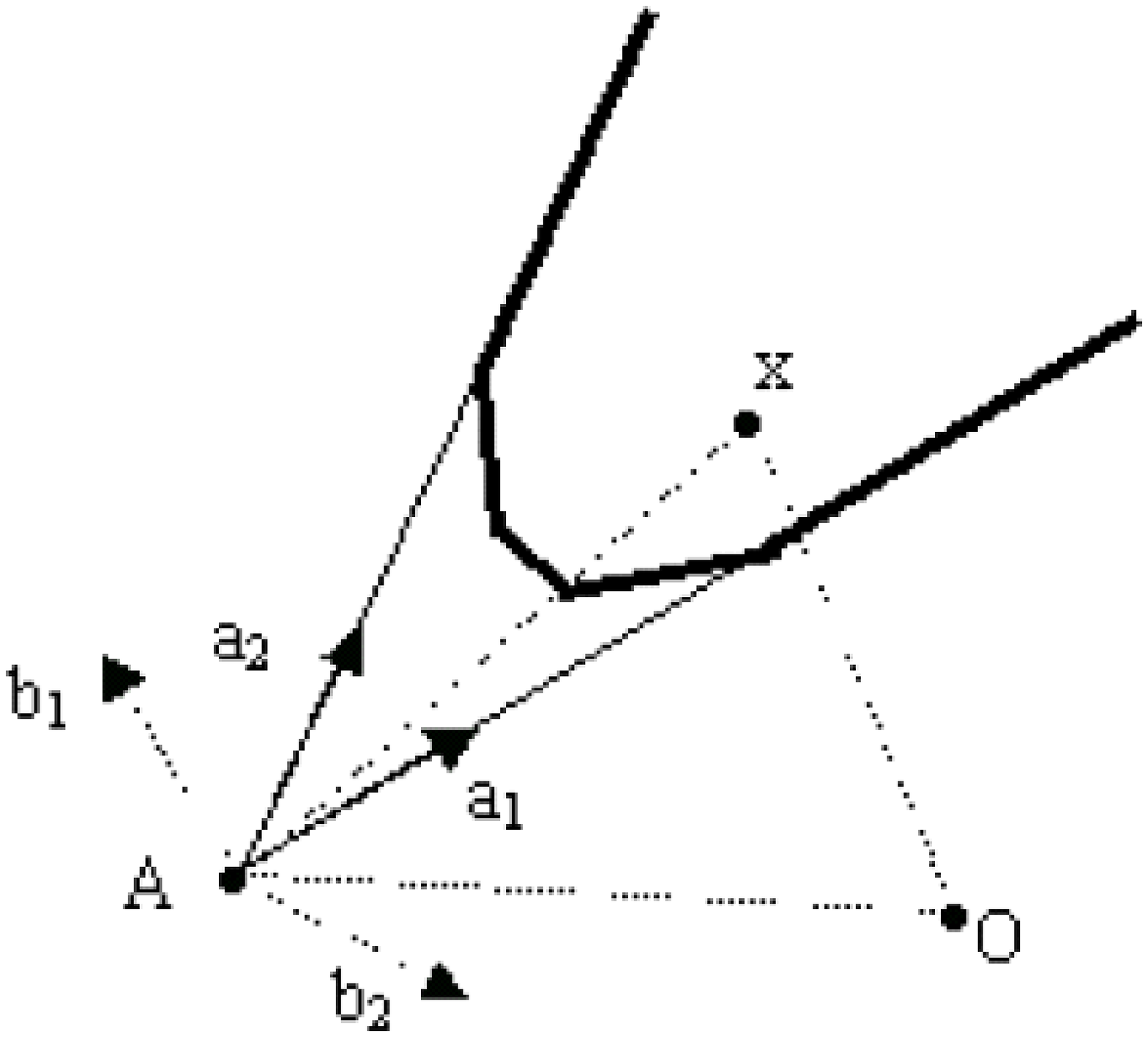}}}
\subfigure[A Cluster within an obtuse angle]{\label{Prop_1b}\scalebox{0.2}{\includegraphics*{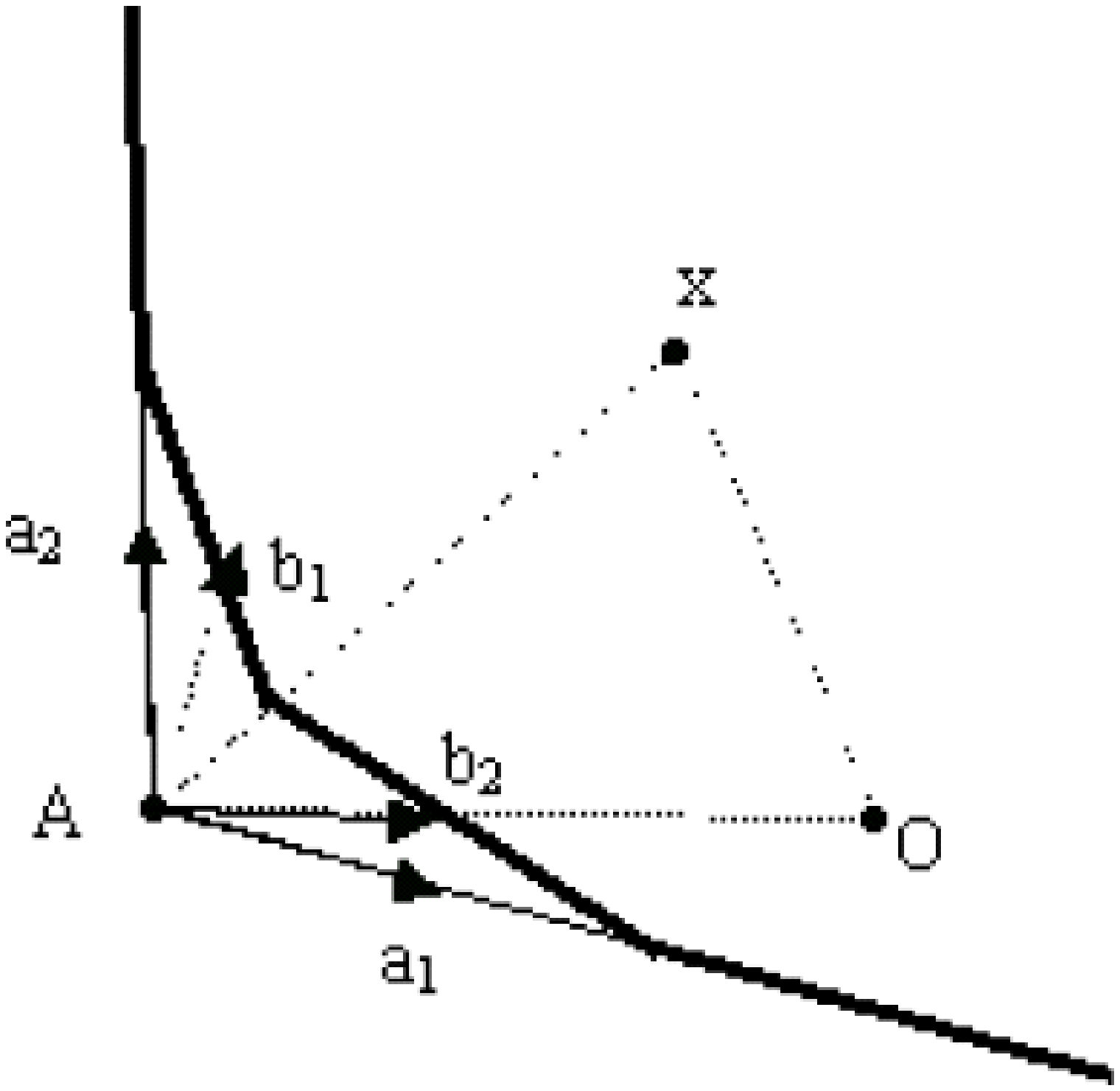}}}
\caption{Notations for the cone containing an unbounded cluster of a SOM; see text for details.}
\label{Prop_1}
\end{center}
\end{figure}

Before going in further details, let us remark that for a SOM with at least 3 classes 
in general position, class $C_0$ covers less than a half plane. Furthermore, we have 
to distinguish two cases for each cluster. First, the cluster may be included in a finite 
compact from $\R^2$. The second case is the case of an infinite cluster i.e. of a cluster which
may does have any neighbour in some direction; this happens to ckusters on the border of the map.

The first case is easely proved. Since $\|X_0\| < R_0$, where $R_0$ can be any constant, 
then we have by triangular inequality:
\begin{equation}
\begin{array}{rcl}
E(\|X_1\|) & < & R_0 + \|Y_0\| \\
 & \leq & R_0 + \max_j(\|Y_j\|).
\end{array}
\end{equation}
As the deformations $Y_j$ are in finite number, the maximum of their norm is finite. 
This proves the first inequality of (\ref{Foster}) in an obvious way for the first case 
(i.e. bounded cluster case). 

The other case thus appens when $\|X_0\| \rightarrow +\infty$. This happens in unbounded 
clusters. The unbounded cluster case is much more technical to prove.

Looking at figure \ref{Prop_1}, we see that each unbounded cluster is included in a cone 
with vertex $A$ and delimited by the normalized vectors $a_1$ and $a_2$. There are 
two possibilities: either $a_1$ and $a_2$ form an acute angle, either an obtuse one, 
as shown in figure \ref{Prop_1a} and figure \ref{Prop_1b} respectively. 

Before going on and applying Foster's criterion, note that the three following 
geometrical properties can be proven:

\subsection*{Property 1.}
Denoting
\begin{equation}\label{prop1}
\lim_{\|x\| \rightarrow \infty} \frac{x}{\|x\|} \cdot a_i = \delta_i,
\end{equation}
we have $\delta_1$ and $\delta_2$ both positive in the acute angle
case, while either $\delta_1$ or $\delta_2$ is positive for an obtuse
angle. Indeed, using the origin $O$, we define:
\begin{equation}
\overrightarrow{Ox} = \overrightarrow{OA} + \overrightarrow{Ax}.
\end{equation}

We thus have:
\begin{equation}
\frac{x}{\|x\|} \cdot a_i = \frac{\overrightarrow{OA} \cdot a_i}{\|x\|}
+ \frac{\overrightarrow{Ax}}{\|\overrightarrow Ax\|}\frac{\|\overrightarrow Ax\|}{\|x\|} \cdot a_i
\end{equation}
which can be bounded by a strictly positive constant as $\frac{\overrightarrow{OA} \cdot a_i}{\|x\|}
\rightarrow 0$ and $\frac{\|\overrightarrow{Ax}\|}{\|x\|} \rightarrow 1$ for 
$\|x\| \rightarrow +\infty$.

\subsection*{Property 2.}
We define $b_1$ such that the angle $(a_1, b_1)$ is $+\frac{\pi}{2}$.
Similarly $b_2$ is defined such that the angle $(b_2, a_2)$ is also $+\frac{\pi}{2}$. 
Then, for both the acute and obtuse angle cases, we have:
\begin{equation}\label{prop2}
\inf_{x \in C} \frac{\overrightarrow{Ax}}{\|x\|} \cdot b_i = r_i > 0,
\end{equation}
where $C$ is the considered cone which has border
vectors $a_1$ and $a_2$.

Rewrite the first term of (\ref{prop2}) as:
\begin{equation}
\inf_{x \in C} \frac{\overrightarrow{Ax}}{\|x\|} \cdot b_i 
= \inf_{x \in C} \frac{\overrightarrow{Ax}}{\|\overrightarrow Ax\|} \frac{\|\overrightarrow Ax\|}{\|x\|} \cdot b_i;
\end{equation}
the result is obtained easily since $\frac{\|\overrightarrow Ax\|}{\|x\|} \rightarrow 1$ when
$\|x\| \rightarrow +\infty$.

\subsection*{Property 3.}
\begin{figure}[!hbt]
\begin{center}
\subfigure[Acute angle case.]{\label{Prop_3a}\scalebox{0.2}{\includegraphics*{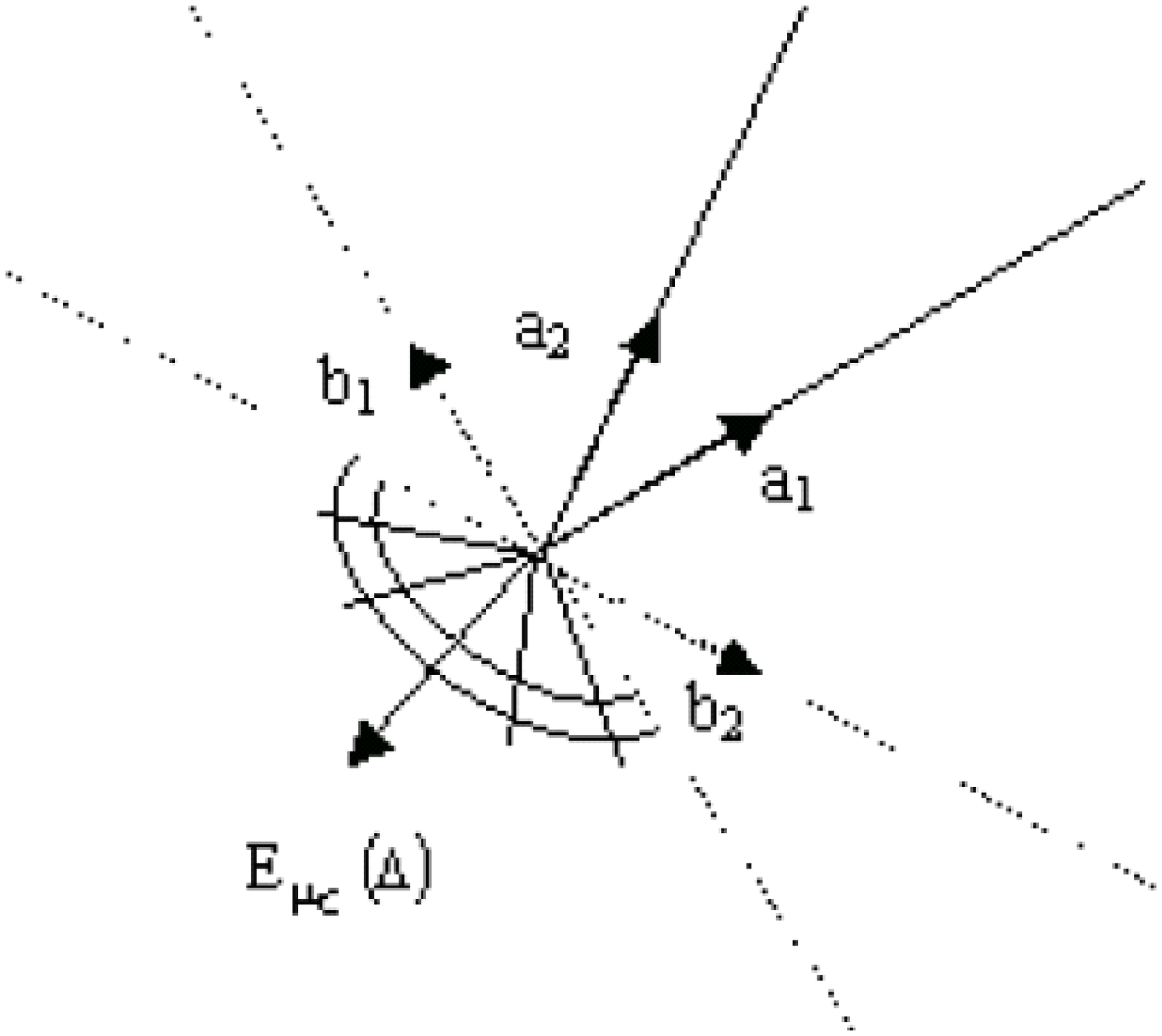}}}
\subfigure[Obtuse angle case.]{\label{Prop_3b}\scalebox{0.2}{\includegraphics*{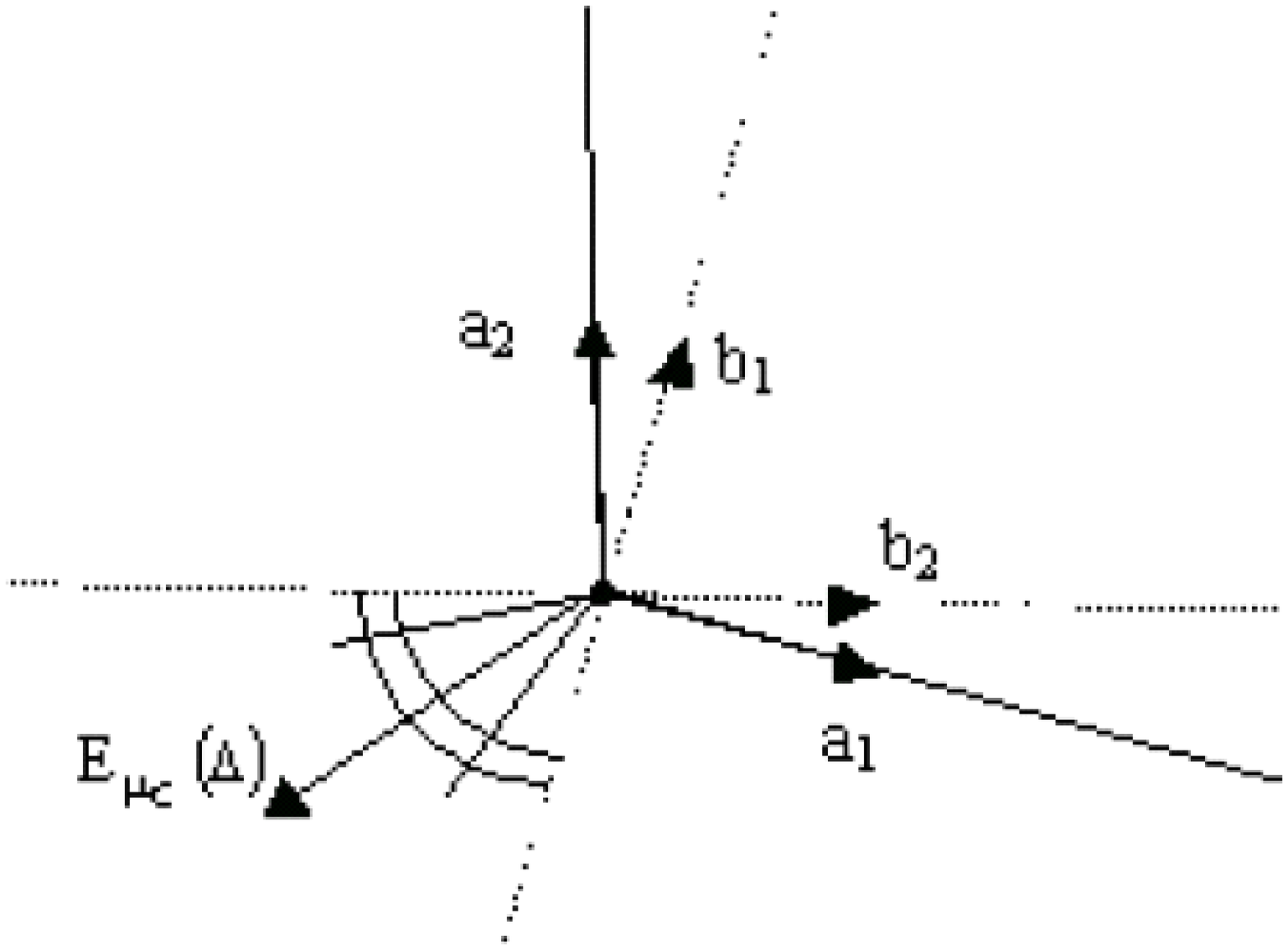}}}
\caption{Third geometrical property, see text for details.}
\label{Prop_3}
\end{center}
\end{figure}

Assume that:
\begin{equation}
E_{\mu_0}(Y_0) \cdot a_1 < 0\ \rm{ and }\ E_{\mu_0}(Y_0) \cdot a_2 < 0
\end{equation}
where $\mu_0$ is the empirical distribution corresponding to class $C_0$ in the transition 
matrix. Denoting
\begin{equation}
E_{\mu_0}(Y_0) \cdot a_i = -\gamma_i < 0
\end{equation}
with $\gamma_i > 0$, then we have:
\begin{equation}
E_{\mu_0}(Y_0) \cdot b_i < 0
\end{equation}
for either $i = 1$ or $i = 2$ in case of an acute angle (figure \ref{Prop_3a}) or for both 
of $i = 1$ and $i = 2$ for the obtuse case (figure \ref{Prop_3b}).

Note that the initial assumption can easily be proved numerically.

Those three properties will be used as lemmas in the following. 
Now we can apply Foster's criterion for the unbounded cluster case.

\subsection*{Foster's criterion}
Considering an unbounded class $C_0$ and the corresponding transition distribution, 
with $g(x) = \|x\|^2$, we have
\begin{equation}\label{dvlp_foster}
\begin{array}{l}
E(g(X_1) | X_0 = x) - g(x) \\
\hspace{0.5cm} = E(g(X_0 + Y_0 | X_0 = x) - g(x) \\
\hspace{0.5cm} = E(\|X_0 + Y_0\|^2 | X_0 = x) - \|x\|^2 \\
\hspace{0.5cm} = 2\|x\| \left[\frac{x \cdot E_{\mu_0}(Y_0)}{\|x\|} + \frac{E_{\mu_0}(\|Y_0\|^2)}{2\|x\|} \right].
\end{array}
\end{equation}

The second term between the brackets can be bounded by a strictly positive constant $\alpha_0$. 
Indeed, as $\|Y_0\|^2$ is finite, $E_{\mu_0}(\|Y_0\|^2) < M_0$ is also finite. Therefore, 
for $\alpha_0 > 0$ and $\|x\| > \frac{M_0}{\alpha_0}$, we have
\begin{equation}\label{dvlp_2nd_term}
\frac{1}{\|x\|} E_{\mu_0}(\|Y_0\|^2) < \alpha_0.
\end{equation}

For the first term, we chose either $i = 1$ or $i = 2$ such that:
\begin{equation}\label{conds}
\left\lbrace
\begin{array}{l}
\displaystyle \lim_{\|x\| \rightarrow +\infty} \frac{x}{\|x\|} \cdot a_i = \delta_i > 0,\\
E_{\mu_0}(Y_0) \cdot b_i < 0.
\end{array}
\right.
\end{equation}

In case of an unbounded cluster, those two conditions are fulfilled using Properties 1. and 3.

By hypothesis, suppose that $i = 2$ satisfies those two conditions (\ref{conds}). 
The term $E_{\mu_0}(Y_0)$ can be decomposed in the $(b_2, a_2)$ basis. Then, for
$\|x\|$ sufficiently large, as:
\begin{itemize}
\item $E_{\mu_0}(Y_0) \cdot a_2 = -\gamma_2$ by Property 3.;
\item $\displaystyle \frac{x}{\|x\|} \cdot a_2 > \frac{\delta_2}{2}$ by Property 1.;
\item $E_{\mu_0}(Y_0) \cdot b_2 < 0$ by Property 3.;
\item $\displaystyle \frac{x}{\|x\|} \cdot b_2 \geq \frac{r_2}{2}$ as 
$\overrightarrow{Ox} = \overrightarrow{OA} + \overrightarrow{Ax}$ and by Property 2.;
\end{itemize}
we have
\begin{equation*}
\begin{array}{l}
\displaystyle \frac{x}{\|x\|} E_{\mu_0}(Y_0)\\
\hspace{0.5cm} 
\leq \underbrace{(E_{\mu_0}(Y_0) \cdot a_2)}_{=-\gamma_2} \underbrace{\left(\frac{x}{\|x\|} \cdot a_2\right)}_{>\frac{\delta_2}{2}}\\
\hspace{1cm}
+ \underbrace{(E_{\mu_0}(Y_0) \cdot b_2)}_{<0} \underbrace{\left(\frac{x}{\|x\|} \cdot b_2\right)}_{\geq \frac{r_2}{2}}\\
\hspace{0.5cm}
\displaystyle < -\gamma_2 \frac{\delta_2}{2},
\end{array}
\end{equation*}
when $\|x\|$ is large enough, denoted here $\|x\| > L_0$.

The same development can be achieved using $i = 1$ to satisfy the two initial conditions 
(\ref{conds}). We obtain:
\begin{equation}
\frac{x}{\|x\|} E_{\mu_0}(Y_0) < -\gamma_1 \frac{\delta_1}{2},
\end{equation}
when $\|x\| > L'_0$.

Equation (\ref{dvlp_foster}) can now be simplified in:
\begin{equation}\label{dvlp_foster_2}
\begin{array}{l}
E(g(X_1) | X_0 = x) - g(x) \\
\hspace{0.5cm}
= 2 \|x\| \left[ \frac{x \cdot E_{\mu_0}(Y_0)}{\|x\|} + \frac{E_{\mu_0}(\|Y_0\|^2)}{2\|x\|}\right] \\
\hspace{0.5cm}
< 2 \|x\| \left[-\alpha_0 + \frac{1}{2} \alpha_0 \right] \\
\hspace{0.5cm}
= -2 \|x\| \frac{\alpha_0}{2},
\end{array}
\end{equation}
where $\|x\| > K_0 = \max(L_0, L'_0)$ and $\alpha_0$ in (\ref{dvlp_2nd_term}) is
chosen such that 
$\alpha_0 = \min \left(\frac{\gamma_1 \delta_1}{2}, \frac{\gamma_2 \delta_2}{2} \right)$.

This development has been done for cluster $C_0$. All values $\alpha_0$, $M_0$, $L_0$, $K_0$ 
depends on this cluster $C_0$. Now considering all unbounded clusters $C_i$ and taking 
$\alpha = \inf_{C_i} \alpha_i$ and $K = \sup_{C_i} K_i$, we have:
\begin{equation}\label{all_classes}
\begin{array}{l}
\forall \|x\| \geq K : \\
\hspace{0.5cm} 
\displaystyle \frac{x E_{\mu_0}(Y_0)}{\|x\|} + \frac{E_{\mu_0}(\|Y_0\|^2)}{2 \|x\|} < - \frac{\alpha}{2} < 0.
\end{array}
\end{equation}

Finally, we obtain, using (\ref{all_classes}) in (\ref{dvlp_foster_2}):
\begin{equation}
E(g(X_1) | X_0 = x) - g(x) < -\alpha \|x\|,
\end{equation}
where the right member tends to $-\infty$ for $\|x\| \rightarrow +\infty$.

To conclude, we define the set $\Omega$ used in Foster's criterion according to
\begin{equation}\label{omega_def}
\displaystyle \Omega = \left(\bigcup_{i \in I} C_i\right) \bigcup \left\{ X_0|\ \|X_0\| < K \right\},
\end{equation}
where $I$ denotes the set of bounded cluster indexes as discussed in the introduction to the
proof. With this definition, the above developments prove Foster's criterion (\ref{Foster}). 
Thus the Markov chain defined by the $X_i$ for $i > 0$ is ergodic, and admits a unique 
stationary distribution.


\begin{thebibliography}{14}
\bibitem{Kohonen95} T. Kohonen, Self-organising Maps, Springer Series in Information Sciences, Vol. 30, Springer, Berlin, 1995.
\bibitem{deBodt04} E. de Bodt, M. Cottrell, P. Letremy, M. Verleysen, On the use of Self-Organizing Maps to accelerate vector quantization, Neurocomputing, Elsevier, Vol. 56 (January 2004), pp. 187-203.
\bibitem{Cottrell98} M. Cottrell, J.-C. Fort, G. Pag\`es, Theoretical aspects of the SOM algorithm, Neurocomputing, 21, p119-138, 1998.
\bibitem{Cottrell97} M. Cottrell, E. de Bodt, M. Verleysen, Kohonen maps versus vector quantization for data analysis, European Symp. on Artificial Neural Networks, April 1997, Bruges (Belgium), D-Facto pub. (Brussels), pp. 187-193.
\bibitem{Cottrell96} M. Cottrell, E. de Bodt, Ph. Gr\'egoire, Simulating Interest Rate Structure Evolution on a Long Term Horizon: A Kohonen Map Application, Proceedings of Neural Networks in The Capital Markets, Californian Institute of Technology, World Scientific Ed., Pasadena, 1996.
\bibitem{Cottrell98b} M. Cottrell, B. Girard, P. Rousset, Forecasting of curves using a Kohonen classification, Journal of Forecasting, Vol. 17, pp. 429-439, 1998.
\bibitem{Walter90} J. Walter, H. Ritter, K. Schulten, Non-linear prediction with self-organising maps, Proc. of IJCNN, San Diego, CA, 589-594, July 1990. 
\bibitem{Vesanto97} J. Vesanto, Using the SOM and Local Models in Time-Series Prediction, In Proceedings of Workshop on Self-Organizing Maps (WSOM'97), Espoo, Finland, pp. 209-214, 1997.
\bibitem{Koskela98} T. Koskela, M. Varsta, J. Heikkonen, and K. Kaski, Recurrent SOM with Local Linear Models in Time Series Prediction, European Symp. on Artificial Neural Networks, April 11 1998, Bruges (Belgium), D-Facto pub. (Brussels), pp. 167-172.
\bibitem{Lendasse98} A. Lendasse, M. Verleysen, E. de Bodt, M. Cottrell, Ph. Gr\'egoire, Forecasting Time-Series by Kohonen Classification, European Symp. on Artificial Neural Networks, April 1998, Bruges (Belgium), D-Facto pub. (Brussels), pp. 221-226.
\bibitem{Verleysen99} M. Verleysen, E. de Bodt, A. Lendasse, Forecasting financial time series through intrinsic dimension estimation and non-linear data projection, in Proc. of International Workconference on Artificial and Natural Neural networks (IWANN'99), Springer-Verlag Lecture Notes in Computer Science, n 1607, pp. II596-II605, June 1999.
\bibitem{Weigend94} A. S. Weigend, N.A. Gershenfeld, Times Series Prediction: Forecasting the future and Understanding the Past, Addison-Wesley Publishing Company, 1994. 
\bibitem{deBodt02}  E. de Bodt, M. Cottrell, M. Verleysen, Statistical tools to assess the reliability of selforganizing maps, Neural Networks, Elsevier, Vol. 15, Nos. 8-9 (October-November 2002), pp. 967-978.
\bibitem{Fayolle95} G. Fayolle, V. A.Malyshev, M. V. Menshikov, Topics in constructive theory of countable Markov chains, Cambridge University Press, 1995.
\end{thebibliography}
\end{document}